\documentclass{article}

\usepackage[final]{neurips_2022}

\usepackage[utf8]{inputenc} 
\usepackage[T1]{fontenc}    
\usepackage{hyperref}       
\usepackage{url}            
\usepackage{booktabs}       
\usepackage{amsfonts}       
\usepackage{nicefrac}       
\usepackage{microtype}      
\usepackage{xcolor}         

\usepackage{wrapfig}
\usepackage{graphicx}
\usepackage{caption}
\usepackage{subcaption}
\usepackage{amssymb}
\usepackage{amsthm}
\usepackage{amsmath}
\usepackage{soul}
\usepackage{bbm}
\usepackage{listings}
\usepackage{tabularx}
\usepackage{enumitem}

\usepackage[T1]{fontenc}
\usepackage[utf8]{inputenc}
\usepackage{authblk}

\graphicspath{{figures/}}

\title{\papertitle}

\author[1,2]{Victor Zhong\thanks{Corresponding author Victor Zhong~\url{vzhong@cs.washington.edu}}}
\author[3]{Jesse Mu}
\author[1,2]{Luke Zettlemoyer}
\author[4,5]{Edward Grefenstette}
\author[4]{Tim Rocktäschel}
\affil[1]{University of Washington}
\affil[2]{Meta AI Research}
\affil[3]{Stanford University}
\affil[4]{University College London}
\affil[5]{Cohere}

\renewcommand\footnotemark{}

\newcommand{\papertitle}{{Improving Policy Learning via Language Dynamics Distillation
}}
\newcommand{\methodname}{{Language Dynamics Distillation}}
\newcommand{\methodnameshort}{{\texttt{LDD}}}

\newcommand{\mdp}{{\mathcal{M}}}

\newcommand{\statespace}{{\mathcal{S}}}
\newcommand{\state}{{s}}
\newcommand{\action}{{a}}
\newcommand{\actionspace}{{\mathcal{A}}}
\newcommand{\transitionprob}{{P}}
\newcommand{\reward}{{r}}
\newcommand{\valuefunction}{{V}}
\newcommand{\policy}{{\pi}}
\newcommand{\discountfactor}{{\gamma}}
\newcommand{\discountreturn}{{G}}
\newcommand{\costfunction}{{J}}
\newcommand{\trajreward}{{\mathcal{R}}}
\newcommand{\demonstrations}{{\mathcal{T}}}
\newcommand{\traj}{{\tau}}
\newcommand{\trajlen}{{T}}
\newcommand{\expect}{{\mathbb{E}}}
\newcommand{\actorcritic}{{\textrm{ac}}}
\newcommand{\expert}{{\sigma}}
\newcommand{\dynamicsmodel}{{\delta}}
\newcommand{\rep}{{\textrm{rep}}}
\newcommand{\modulerep}{{f_\rep}}
\newcommand{\moduledynamics}{{f_\dynamicsmodel}}
\newcommand{\modulepolicy}{{f_\policy}}
\newcommand{\modulevalue}{{f_\valuefunction}}
\newcommand{\genericparam}{{\theta}}
\newcommand{\policyparam}{{\theta}}
\newcommand{\valueparam}{{\phi}}
\newcommand{\dynamicsparam}{{\zeta}}
\newcommand{\distill}{{d}}
\newcommand{\similarity}{{\mathit{sim}}}
\newcommand{\teacher}[1]{{\tilde{#1}}}
\newcommand{\norm}[1]{{|{#1}|}}

\begin{document}

\maketitle

\begin{abstract}
Recent work has shown that augmenting environments with language descriptions improves policy learning.
However, for environments with complex language abstractions, learning how to ground language to observations is difficult due to sparse, delayed rewards.
We propose~\methodname~(\methodnameshort), which pretrains a model to predict environment dynamics given demonstrations with language descriptions, and then fine-tunes these language-aware pretrained representations via reinforcement learning (RL).
In this way, the model is trained to both maximize expected reward and retain knowledge about how language relates to environment dynamics.
On SILG, a benchmark of five tasks with language descriptions that evaluate distinct generalization challenges on unseen environments (NetHack, ALFWorld, RTFM, Messenger, and Touchdown),~\methodnameshort~outperforms tabula-rasa RL, VAE pretraining, and methods that learn from unlabeled demonstrations in inverse RL and reward shaping with pretrained experts.
In our analyses, we show that language descriptions in demonstrations improve sample-efficiency and generalization across environments, and that dynamics modeling with expert demonstrations is more effective than with non-experts.
\end{abstract}

\section{Introduction}
Language is a powerful medium that humans use to reason about abstractions---its compositionality allows efficient descriptions that generalize across environments and tasks.
Consider an agent that follows instructions to clean the house (e.g.~find the dirty dishes and wash them).
In tabula-rasa reinforcement learning (RL), the agent observes raw perceptual features of the environment, then grounds these visual features to language cues to learn how to behave through trial and error.
In contrast, we can provide the agent with language descriptions that describe abstractions which are present in the environment (e.g.~\emph{there is a sink to your left and dishes on a table to your right}), thereby simplifying the grounding challenge.
Language descriptions of observations occur naturally in many environments such as text prompts in graphical user interfaces~\citep{Zheran2018RL4Web}, dialogue~\citep{he2018decoupling}, and interactive games~\citep{kuttler2020nethack}.
Recent work has also shown improvements in visual manipulation~\citep{shridhar2021alfworld} and navigation~\citep{zhong2021silg,tam2022semanticExploration} by captioning the observations with language descriptions.
Despite these gains, learning how to interpret language descriptions is difficult through RL, especially on environments with complex language abstractions~\citep{zhong2021silg}.

\begin{figure}
    \centering
    \includegraphics[width=\linewidth]{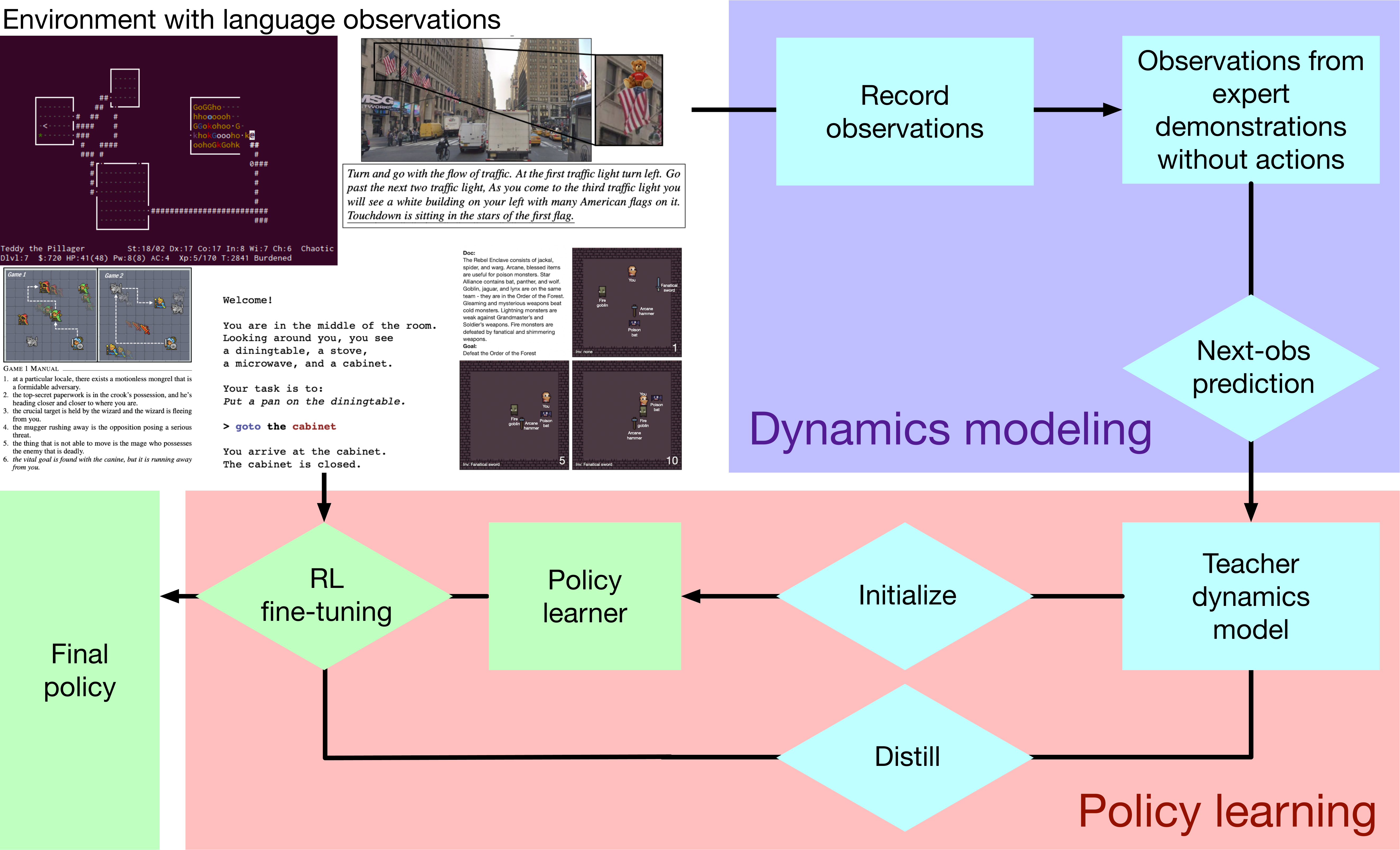}
    \caption{\methodname~(\methodnameshort). \methodnameshort~uses cheap unlabeled demonstrations to learn a dynamics model of the environment, which is used to initialize and distill grounded representations into the policy learner. During the dynamics modeling phase (purple), we train a teacher model to predict the next observation given prior observations using unlabeled demonstrations. In the policy learning phase (red), we initialize a model with the teacher and distill intermediate representations from the teacher during reinforcement learning. The traditional policy learning loop is shown in green. \methodnameshort-specific components are shown in blue.}
    \label{fig:framework}
\end{figure}

We present~\methodname~(\methodnameshort), a method that improves RL by learning a dynamics model on cheaply obtained unlabeled (i.e.~no action labels) demonstrations with language descriptions.
When learning how to use language descriptions effectively, one central challenge is how to disentangle language understanding from policy performance from sparse, delayed rewards.
Our motivation is to learn initial language grounding via dynamics modeling from an offline dataset, away from the credit assignment and non-stationarity challenges posed by RL.
While labeled demonstrations that tell the agent how to to act in each situation are expensive to collect, for many environments one can cheaply obtain unlabeled demonstrations (e.g.~videos of experts performing the task)~\citep{yang2019imitation,stadie17thirdpersonimitation}.
Intuitively,~\methodnameshort~exploits these unlabeled demonstrations to learn how to associate language descriptions with abstractions in the environment.
This knowledge is then used to bootstrap and more quickly learn policies that generalize to instructions and manuals in new environments.
Given unlabeled demonstrations with language descriptions (e.g.~captions of scene content), we first pretrain the model to predict the next observation given prior observations, similar to language modeling.
A copy of this model is stored as a fixed teacher that grounds language descriptions to predict environment dynamics.
We then train a model with RL, while distilling intermediate representations from the teacher to avoid catastrophic forgetting of how to interpret language descriptions for dynamics modeling.
In this way, the model learns to both maximize expected reward while retaining knowledge about how language descriptions relate to environment dynamics.

We evaluate~\methodnameshort~on the recent SILG benchmark~\citep{zhong2021silg}, which consists of five diverse environments with language descriptions including NetHack~\citep{kuttler2020nethack}, ALFWorld~\citep{shridhar2021alfworld}, RTFM~\citep{zhong2020rtfm}, Messenger~\citep{hanjie2021grounding}, and Touchdown~\citep{chen2018touchdown}.
These environments present unique challenges in language-grounded policy-learning across complexity of instructions, visual observations, action space, reasoning procedure, and generalization.
By learning a dynamics model from cheaply obtained unlabeled demonstrations,~\methodnameshort~consistently outperforms reinforcement learning with language descriptions both in terms of sample efficiency and generalization performance.
Moreover, we compare~\methodnameshort~to other techniques that inject prior knowledge in VAE pretraining~\citep{kingma2013VAE}, inverse reinforcement learning~\citep{hanna2017Grounded,torabi2018behavioral,guo2019hybridRL}, and reward shaping with a pretrained expert~\citep{merel2017adversarialImitation}.
\methodnameshort~achieves top performance on all environments in terms of task completion and reward.
In addition to comparing~\methodnameshort~to other methods, we ablate~\methodnameshort~to quantify the effect of language observations in dynamics modeling, and the importance of dynamics modeling with expert demonstrations.
On two environments where we can control for the presence of language descriptions (NetHack game messages and Touchdown panorama captions), we show that language descriptions improve sample-efficiency and generalization.
Finally, across all environments, we find that dynamics modeling with expert demonstrations is more effective than with non-expert rollouts.

\section{Related Work}

\paragraph{Learning by observing language.}
Recent work studies generalization to language instructions and manuals that specify new tasks and environments.
These settings range from photorealistic/3D navigation~\citep{mattersim,chen2018touchdown,rxr,shridhar2020alfred} to multi-hop reference games~\citep{narasimhan2015language,zhong2020rtfm,hanjie2021grounding}.
We use a collection of these tasks to evaluate~\methodnameshort.
There is also work where understanding language is not necessary to achieve the task, however its inclusion (e.g.~via captions, scene descriptions) makes learning more efficient.
\citet{shridhar2021alfworld}~show that one can quickly learn policies in a simulated kitchen environment described in text, then transfer this policy to the 3D visual environment.
\citet{zhong2021silg}~similarly transform photorealistic navigation to a symbolic form via image segmentation, then learn a policy that transfers to the original photorealistic setting.
In work concurrent to ours,~\citet{tam2022semanticExploration}~generate oracle captions of observations for simulated robotic control and city navigation, which improve policy learning.
\methodnameshort~is complementary to these---in addition to incorporating language descriptions as features, we show that learning a dynamics model from unlabeled demonstrations with language descriptions improves sample efficiency and results in better policies.

\paragraph{Imitation learning from observations.}
There is prior work on model-free as well as model-based imitation learning from observations.
Model-free methods encourage the imitator to produce state distributions similar to those produced by the demonstrator, for example via generative adversarial learning~\citep{merel2017adversarialImitation} and reward shaping~\citep{kimura2018internal}.
In contrast, \methodnameshort~only requires intermediate representations extracted from an expert dynamics model on states encountered by the learner, which are cheaper to compute than rollouts from an expert policy.
Model-based approaches learn dynamics models that predict state-transitions given the current state and an action.
\citet{hanna2017Grounded}~learn an inverse model to map state-transitions to actions, which is then used to annotate unlabeled trajectories for imitation learning.
\citet{edwards2019imitating}~learn a forward dynamics model that predicts future states given state and latent action pairs.
In contrast,~\methodnameshort~does not assume priors over the action space distribution.
For instance, on ALFWorld, our method works even though it is impossible to enumerate the action space.
In our experiments, we extend model-free reward shaping and model-based inverse dynamics modeling to account for language descriptions and compare~\methodnameshort~to these methods.


\paragraph{Representation learning in RL.}
In representation learning for RL, the agent learns representations of the environment using rewards and objectives based on the difference between the state and prior states~\citep{Strehl2008AnAO}, raw visual observations~\citep{jaderberg2017unreal}, learned agent representations~\citep{Raileanu2020RIDE}, and random network observations~\citep{burda2018rnd}.
In intrinsic exploration methods~\citep{Raileanu2020RIDE,burda2018rnd}, the training objective encourages dissimilarity (e.g.~in observation/state space) to prior agent experience so that the agent discovers novel states.
Unlike intrinsic exploration, the distillation objective in~\methodname~encourages similarity to expert behaviour, as opposed to dissimilarity to prior agent experience.
In reconstruction based representation learning methods~\citep{Strehl2008AnAO,jaderberg2017unreal}, the training objective encourages the agent to learn intermediate representations that also capture the dynamics and structure of the environment by reconstructing the observations (e.g.~predicting what objects are in scene).
\methodname~is similar to reconstruction methods for representation learning, however unlike the latter, the dynamics model in~\methodnameshort~is trained on trajectories obtained from an expert policy as opposed to the agent policy.
\methodname~is complementary to intrinsic exploration methods and to reconstruction based representation learning methods.

\section{\methodname}

Recent work improves policy learning by augmenting environment observations with language descriptions~\citep{shridhar2021alfworld,zhong2021silg,tam2022semanticExploration}.
For environments with complex language abstractions, however, learning how to associate language to environment observations is difficult through RL due to sparse, delayed rewards.
In~\methodname~(\methodnameshort), we pretrain the model on unlabeled demonstrations (i.e.~no annotated actions) with language descriptions to predict the dynamics of the environment, then fine-tune the language-aware model via RL.
\methodnameshort~consists of two phases.
In the first dynamics modeling phase, we pretrain the model to predict future observations given unannotated demonstrations.
We store a copy of the model as a fixed teacher that has learned grounded representations useful for predicting how the environment behaves under an expert policy.
In the second reinforcement learning phase, we fine-tune the model through policy learning, while distilling representations from the teacher.
This way, the model is trained to both maximize expected reward and retain knowledge about the dynamics of the environment.
Fig~\ref{fig:framework}~illustrates the components of~\methodnameshort.

\subsection{Background}

\paragraph{Markov decision process.}
Consider a MDP $\mdp = \{\statespace, \actionspace, \transitionprob, \reward, \gamma\}$.
Here, $\statespace$ and $\actionspace$ respectively are the discrete state (e.g.~language goals, descriptions, visual observations) and action spaces of the problem.
$\transitionprob(\state_{t+1} | \state_t, \action_t)$ is the transition probability of transitioning into state $\state_{t+1}$ by taking action $\action_t$ from state $\state_t$.
$\reward(\state, \action)$ is the reward function given some state and action pair.
$\discountfactor$ is a discount factor to prioritize short-term rewards.

\paragraph{Actor-critic methods for policy learning.}
In RL, we learn a policy $\policy (\state ; \policyparam)$ that maps from observations to actions $\policy: \statespace \rightarrow \actionspace$.
Let $\trajreward(\traj)$ denote the total discounted reward over the trajectory $\traj$.
The objective is to maximize the expected reward $\costfunction_\policy(\policyparam) = \expect_\policy[\trajreward(\traj)]$ following the policy $\policy$ by optimizing its parameters $\policyparam$.
For trajectory length $\trajlen$, the policy gradient is
\begin{eqnarray}
\nabla \expect_\policy[\trajreward(\traj)] &=& \expect_\policy \left[\left( \trajreward(\traj) \sum_{t=1}^\trajlen \nabla \log \pi (\action_t, \state_t) \right) \right]
= \expect_\policy \left[\left( \sum_{t=1}^\trajlen \discountreturn_t \nabla \log \pi (\action_t, \state_t) \right) \right]
\end{eqnarray}
where $\discountreturn_t = \sum_{k=0}^\infty \discountfactor^k \reward_{t+k+1}$ is the return or discounted future reward at time $t$.
We consider the actor-critic family of policy gradient methods, where a critic is learned to reduce variance in the gradient estimate.
Let $\valuefunction(\state) = \expect_\policy [\discountreturn_t | \state_t = \state]$ denote the state value, which corresponds to the expected returns by following the policy $\policy$ from a state $\state$.
Actor critic methods estimate the state value function by learning another parametrized function $\valuefunction$ to bootstrap the estimation of the discounted return $\discountreturn_t$.
For instance, with one-step bootstrapping, we have
$\discountreturn_t \approx \reward_{t+1} + \discountfactor \valuefunction(\state_{t+1}; \valueparam)$.
The critic objective is then $\costfunction_\valuefunction(\valueparam) = \frac{1}{2} \left(\reward_{t+1} + \discountfactor \valuefunction(\state_{t+1}; \valueparam) - \valuefunction(\state_t; \valueparam)\right)^2$
We minimize a weighted sum of the policy objective and the critic objective
$\costfunction_\actorcritic(\policyparam, \valueparam) = - \costfunction_\policy(\policyparam) + \alpha_\valuefunction \costfunction_\valuefunction(\valueparam)$.

\subsection{Dynamics modeling during pretraining}
\label{sec:dynamics_modeling}
In addition to policy learning,~\methodname~learns a dynamics model from unlabeled demonstrations to initialize and distill into the policy learner.
Consider a set of demonstrations without labeled actions $\demonstrations_\expert = \{\traj_1, \traj_2, \ldots \traj_n\}$ obtained by rolling out some policy $\expert(\action_t, \state_t)$, where each demonstration $\traj = [\state_1, \state_2, \ldots \state_\trajlen]$ consists of a sequence of observations.
We learn a dynamics model $\dynamicsmodel(\state_1 \ldots \state_t; \dynamicsparam)$ to predict the next observation $\state_{t+1}$ given the previous observations.
\begin{eqnarray}
\label{eqn:similarity}
\costfunction_\dynamicsmodel(\dynamicsparam) &=& \frac{1}{n T} \left( \sum_{i=1}^n \left( \sum_{t=1}^\trajlen \similarity \left( \state_{t+1}, \dynamicsmodel(\state_1, \ldots \state_t; \dynamicsparam) \right) \right) \right)
\end{eqnarray}
where $\similarity$ is a differentiable similarity function between the predicted state $\dynamicsmodel(\state_1, \ldots \state_t)$ and the observed state $\state_{t+1}$, and $\dynamicsparam$ are parameters of the dynamics model.
In the environments we consider, $\similarity$ is the cross-entropy loss across a grid of symbols denoting entities present in the scene.

\subsection{Dynamics distillation during policy learning}
\label{sec:distillation}
Fig~\ref{fig:framework}~shows the decomposition of the model into a representation network $\modulerep$, a policy head $\modulepolicy$, a value head $\modulevalue$, and a dynamics head $\moduledynamics$.
The three heads share parameters because their inputs are formed by the same representation network.
\begin{eqnarray}
\policy(\state_t) &=& \modulepolicy \left(\modulerep \left(\state_t ; \genericparam_\rep \right); \genericparam_\policy \right)\\
\valuefunction(\state_t) &=& \modulevalue \left(\modulerep \left(\state_t ; \genericparam_\rep \right); \genericparam_\valuefunction \right)\\
\dynamicsmodel(\state_t) &=& \moduledynamics \left(\modulerep \left(\state_t ; \genericparam_\rep \right); \genericparam_\dynamicsmodel \right)
\end{eqnarray}
In the first phase of dynamics modeling, we pretrain the model to predict future observations given demonstrations by optimizing $\costfunction_\dynamicsmodel$.
We then store a copy of the model as a fixed teacher $\teacher{\dynamicsmodel}(\state_t) = \teacher{\moduledynamics} \left( \teacher{\modulerep} \left(\state_t ; \teacher{\genericparam}_\rep \right) \right)$.
Let $\norm{X - Y}$ denote the L2 distance between $X$ and $Y$.
During the second phase, in addition to policy learning, we optimize a distillation objective
$
\costfunction_\distill(\genericparam_\rep, \genericparam_\policy, \genericparam_\valuefunction) =
    \norm{\teacher{\modulerep} \left( \state_t ; \teacher{\genericparam}_\rep \right) - \modulerep \left( \state_t ; \genericparam_\rep \right)}
$
to avoid catastrophic forgetting of how to interpret language descriptions for dynamics modeling.
This quantity is the similarity between the feature representation produced by the fixed teacher (e.g.~the pretrained dynamics model) and the feature representation produced by the model.
Because $\teacher{\dynamicsmodel}$ is frozen, the parameters $\teacher{\genericparam}_\rep$ are not included in the objective function $\costfunction_\distill$.
The joint loss for~\methodname~is then
\begin{eqnarray}
\costfunction(\genericparam_\rep, \genericparam_\policy, \genericparam_\valuefunction) &=&
    - \costfunction_\policy(\policyparam)
    + \alpha_\valuefunction \costfunction_\valuefunction(\valueparam)
    + \alpha_\distill \costfunction_\distill(\genericparam_\rep, \genericparam_\policy, \genericparam_\valuefunction)
\end{eqnarray}
To summarize, using unlabeled demonstrations with language descriptions,~\methodnameshort~learns a dynamics model of the environment that grounds language descriptions to environment observations (Section~\ref{sec:dynamics_modeling}).
This prior knowledge is then injected into reinforcement learning via initialization and distillation (Section~\ref{sec:distillation}).

\section{Experiments}
We evaluate~\methodname~on the Situated Interactive Language Grounding benchmark (SILG)~\citep{zhong2021silg}.
SILG consists of five different language grounding environments with diverse challenges in term of complexity of observation space, action space, language, and reasoning procedure.
In all environments, a situated agent observes symbolically (RTFM, Messenger, Nethack, Touchdown) or prose (ALFWorld) rendered visuals and interacts with the environment to follow some instance-specific language goals (e.g.~what to do).
In addition, the agent observes text manuals describing instance-agnostic environment rules (e.g.~entity-role associations).
The learning challenge is to learn a reading agent that generalizes to new environments with different environment rules (e.g.~new entity-team associations, new parts of the map).
The five different environments are as follows.

\subsection{Environments}

\begin{figure}[t]
    \centering
    \includegraphics[width=\linewidth]{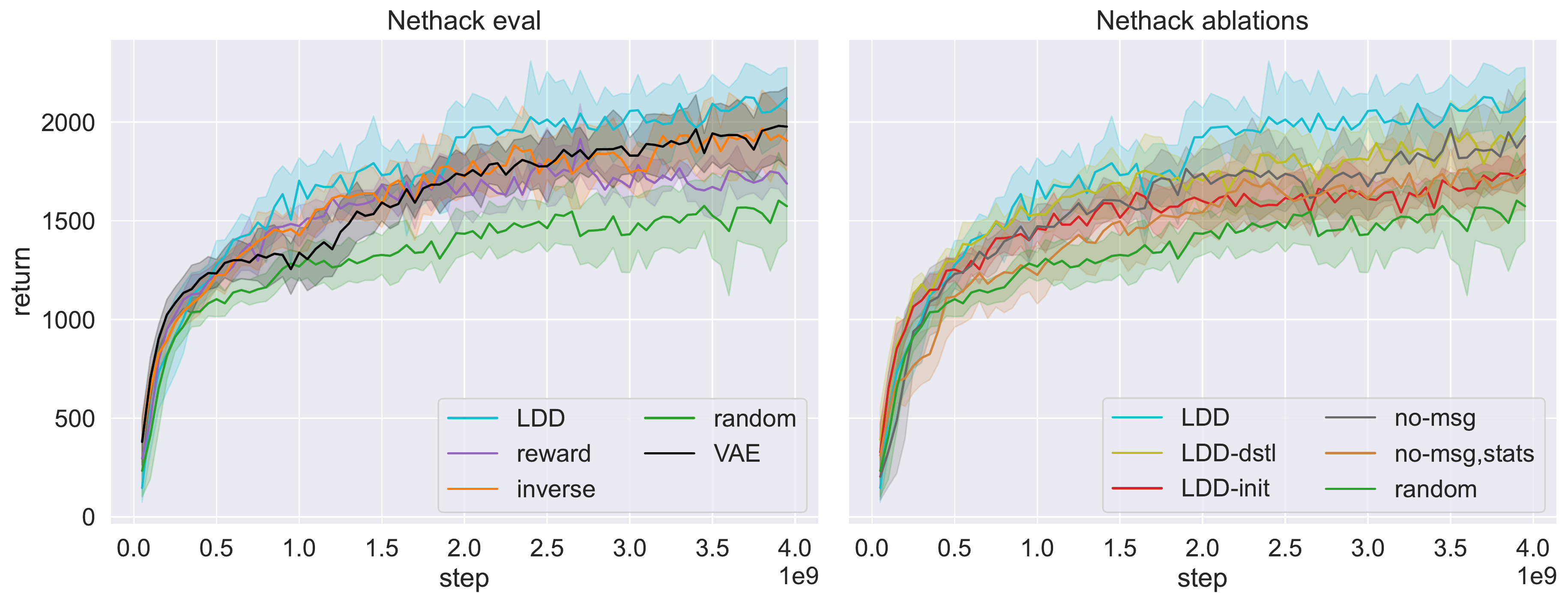}
    \caption{NetHack Challenge comparisons (left) and ablations (right). LDD consistently outperforms other methods.}
    \label{fig:nethack}
\end{figure}

\paragraph{NetHack}
\citep{kuttler2020nethack}:
The agent must descend a procedurally generated dungeon.
Its primarily challenge is in large state space and partial observability, as the map remains obscured until exploration.
Observations include a symbolic grid of entity IDs, in-game message, and description of character stats.
The agent chooses among a fixed set of actions such as movement, picking up/buying/selling items, and attacking.
We evaluate not on SILGNethack but the full NetHack challenge, a difficult game for humans with $\sim$15\% expert win rate~\citep{street2022ObservationsOnExpertNethack}.
\paragraph{SymTouchdown}
\citep{zhong2021silg}:
A symbolic version of Touchdown~\citep{chen2018touchdown} where the agent navigates segmentation maps of Street View panoramas following long instructions.
The primary challenge is reading long, natural instructions that describe photorealistic images.
Evaluation is on new navigation instructions.
Observations include a grid of segmentation class IDs corresponding to a discretized Google Street View panorama, synthetic captions of where objects are relative to the agent (e.g.~\emph{to your right you see a lot of road and some cars}),  and language navigation instructions.
The agent selects from a list of radial directions to proceed to the next panorama.
\paragraph{ALFWorld}
\citep{shridhar2021alfworld}:
The agent navigates and manipulates objects inside a kitchen which is described via textual descriptions.
ALFWorld is challenging due to its large (>50) text action space that vary across scenes.
Evaluation is on unseen instructions.
Observations include a textual description of the scene and language goals (e.g.~\emph{put a clean sponge on the metal rack}).
The agent chooses among a variable set of language actions (e.g.~\emph{open drawer 1}).
\paragraph{RTFM}
\citep{zhong2020rtfm}:
The agent interprets a game manual and instruction to acquire the correct items to fight the correct monsters.
Its main challenge is in multi-step reasoning that combines world observations with texts describing multiple entities.
Evaluation is on a set of manuals distinct from those in training; hence the agent cannot memorize training manuals and must learn to read correctly in order to generalize.
Observations include a symbolic grid containing names of entities present, manual describing high level game rules, agent inventory, and the language instruction.
The agent chooses among a fixed set of movements.
We train and evaluate on the first curriculum stage.
\paragraph{Messenger}
\citep{hanjie2021grounding}:
The agent delivers a message from a source entity to a target while avoiding an enemy.
The entities are referred to in text by many names, which have no lexical overlap with their symbol ID, hence the core challenge is in mapping language entity references in text to observed symbolic entity IDs.
Evaluations are on new entity-role assignments (e.g.~who carries the message).
Observations include a symbolic grid containing symbol IDs of entities present, and a manual of entities and roles.
The agent chooses among a fixed set of movements.
We train and evaluate on the second curriculum stage.
\begin{figure}[t]
    \centering
    \includegraphics[width=\linewidth]{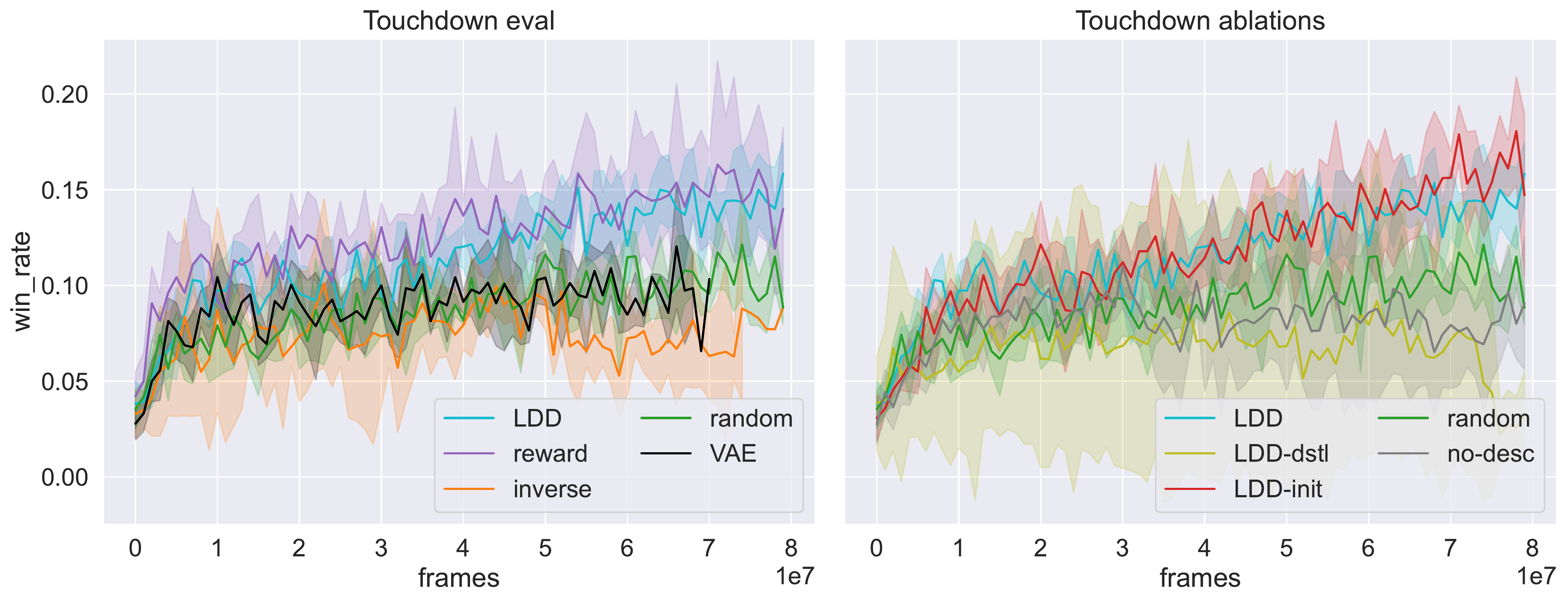}
    \caption{Touchdown comparisons (left) and ablations (right). Methods that distill or reward shape (\methodnameshort, \texttt{reward}, \texttt{LDD-init}) outperform those that do not.}
    \label{fig:touchdown}
\end{figure}

\begin{figure}[t]
    \centering
    \includegraphics[width=\linewidth]{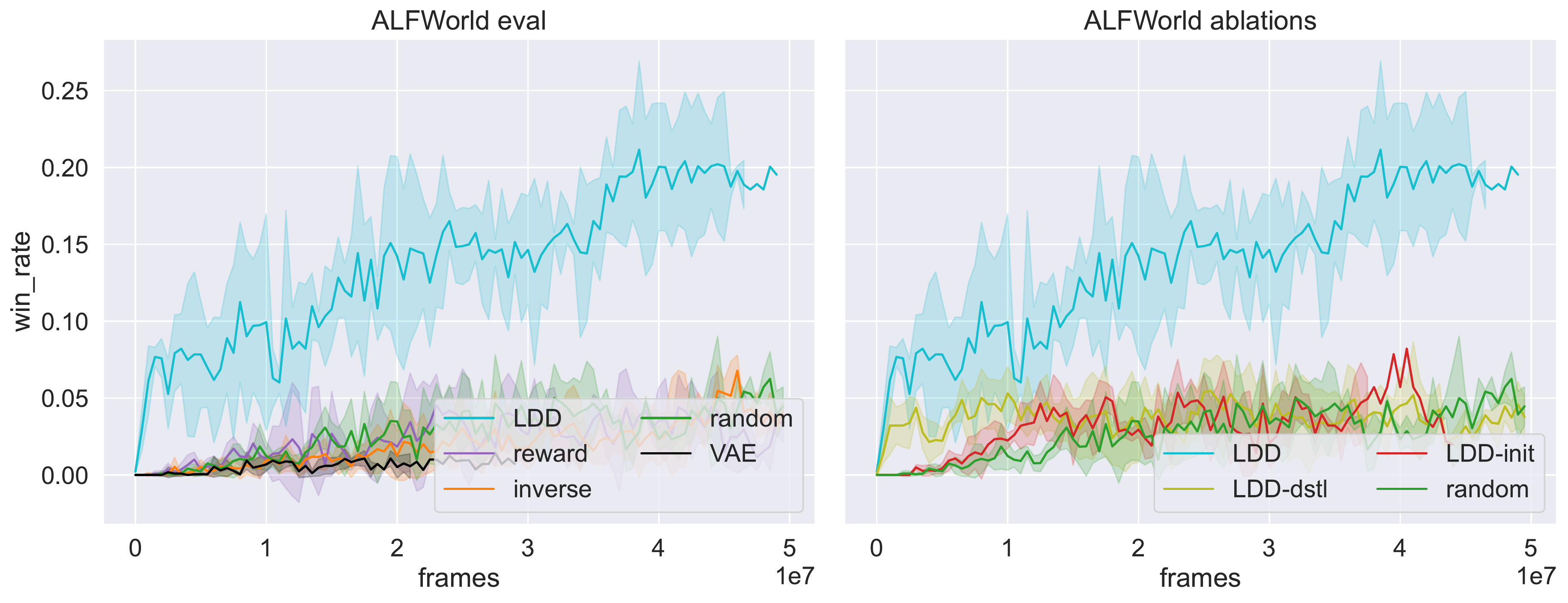}
    \caption{ALFWorld comparisons (left) and ablations (right).~\methodnameshort~consistently outperforms other methods.}
    \label{fig:alfworld}
\end{figure}

\begin{figure}[t]
    \centering
    \includegraphics[width=\linewidth]{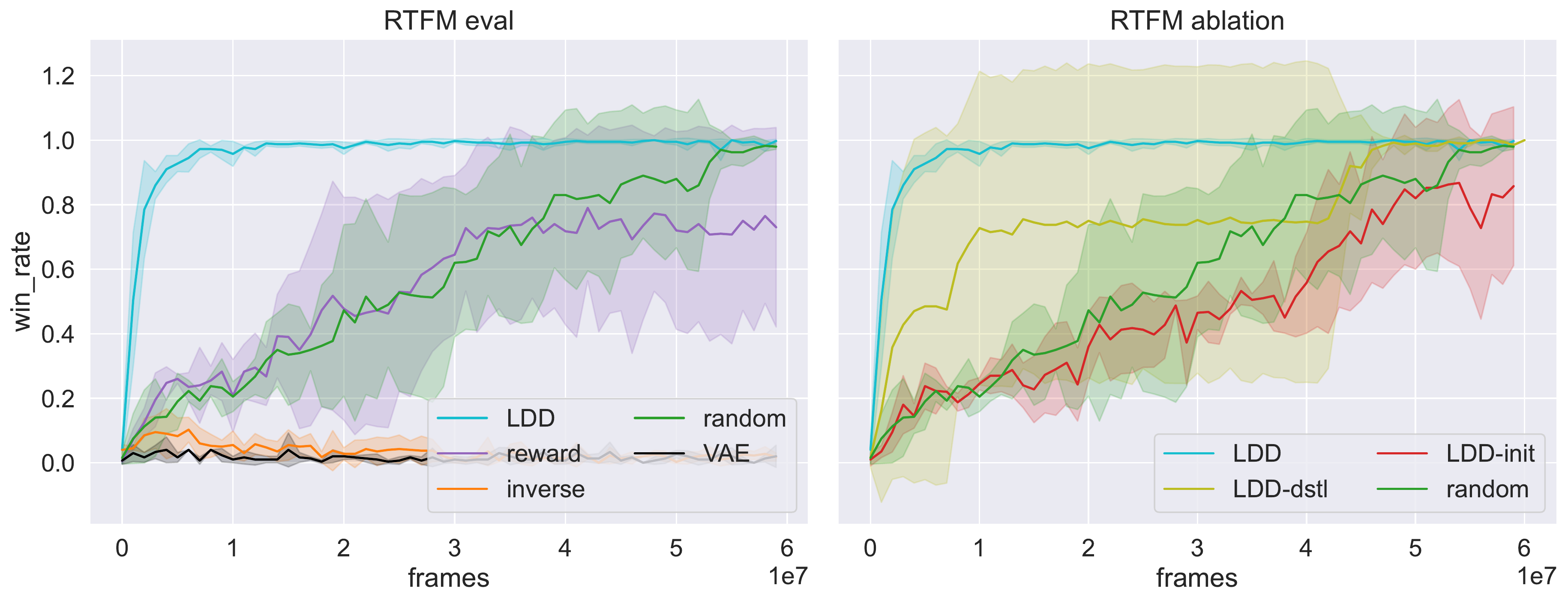}
    \caption{RTFM comparisons (left) and ablations (right).~\methodnameshort~consistently outperforms other methods. Because of the multi-step reasoning nature of RTFM solutions, partially complete strategies (e.g.~able to do 2/3/4 cross-references) result in step-wise gains in win-rates. Strategies at different levels of completion result in larger variances when averaged.}
    \label{fig:rtfm}
\end{figure}

\begin{figure}[t]
    \centering
    \includegraphics[width=\linewidth]{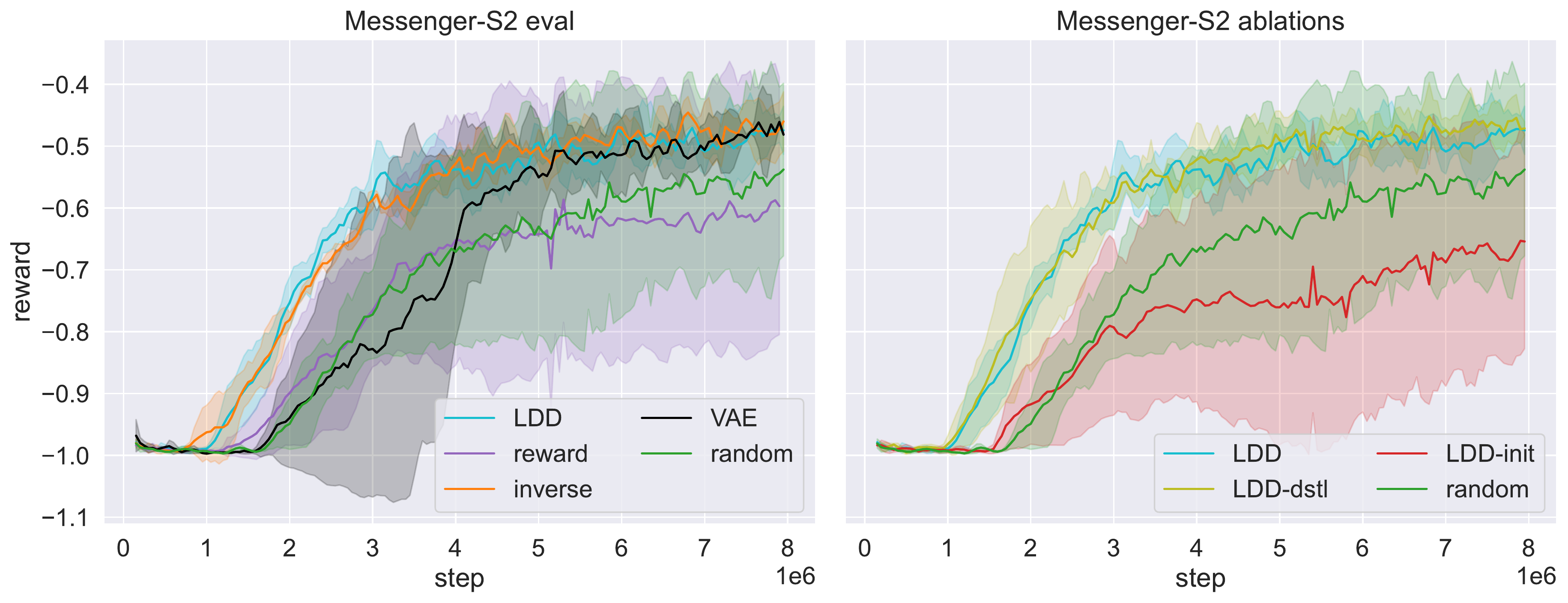}
    \caption{Messenger comparisons (left) and ablations (right). Methods that pretrain to initialize (\methodnameshort, \texttt{inverse}, \texttt{LDD-distill}) outperform those that do not.}
    \label{fig:messenger}
\end{figure}

\subsection{Method and Baselines}
\paragraph{Reinforcement learning with language descriptions from scratch.}
We train a base tabula-rasa policy learner from random initialization.
For NetHack, we train the base policy learner from~\citep{kuttler2020nethack}~using~\texttt{moolib}~\citep{mella2022moolib}.
For RTFM, ALFWorld, and Touchdown, we train the SIR model from~\citep{zhong2021silg}~using Torchbeast~\citep{kuttler2019torchbeast,espeholt2018impala}.
For Messenger, we train the EMMA model from~\citet{hanjie2021grounding} using PPO~\citep{schulman2017ppo}.
For NetHack, we train the base policy learner from~\citep{kuttler2020nethack}~using~\texttt{moolib}~\citep{mella2022moolib}.
For RTFM, ALFWorld, and Touchdown, we train the SIR model from~\citep{zhong2021silg}~using Torchbeast~\citep{kuttler2019torchbeast,espeholt2018impala}.
For Messenger, we train the EMMA model from~\citet{hanjie2021grounding} using PPO~\citep{schulman2017ppo}.
\paragraph{Pretraining representations via a variational autoencoder.}
We pretrain a variational autoencoder (VAE), a common approach for representation learning, that predicts the intermediate representation just before the policy head~\citep{kingma2013VAE}.
This VAE has the same architecture as the policy learner, and is used to initialize the policy learner.
The training procedure for the VAE is as described in~\citet{ha2018worldmodels}.

\paragraph{\methodname~(\methodnameshort)}
We train~\methodnameshort~variants of the baseline policy learners for each environment, where we pretrain the model to perform dynamics modeling on unannotated demonstrations.
For NetHack, we use 100k screen-recordings (where actions are not annotated and cannot be trivially reverse engineered due to ambiguity in observations) of human-playthroughs from the alt.org NetHack public server.
For Touchdown, we use unannotated demonstrations by human players.
For ALFWorld, we use trajectories obtained from an A* planner with full state and goal knowledge, with actions removed.
For RTFM and Messenger, we train expert trajectories until convergence, and sample 10k rollouts from the experts from which we remove action labels.
\paragraph{Reward shaping with expert.}
Methods such as~\citet{merel2017adversarialImitation} reward shape with an expert by encouraging the agent to produce states similar to a demonstrator.
To compare~\methodnameshort~to this idea, we use the dynamics model to predict the next observation under expert policy.
The difference (e.g.~accuracy in symbol prediction across grid) between the predicted observation and the actual observation after taking the action proposed by the agent is used as a penalty (e.g.~negative auxiliary reward).
This method is listing as~\texttt{reward} in experiment figures and uses the same unlabeled demonstrations as~\methodnameshort.
%
\paragraph{Inverse reinforcement learning.}
Another class of methods learn a inverse dynamics model with which infer actions in unlabeled demonstrations, then learn to imitate the pseudo-labeled demonstrations~\citep{hanna2017Grounded,torabi2018behavioral}.
To compare to this method, we first train the base policy learner for 10k episodes, then collect 10k rollouts to train an inverse dynamics model that predicts current action given current and future observations.
This inverse model is used to annotate the original unlabeled demonstrations for imitation learning~\citep{torabi2018behavioral}.
Because these imitation policies do not generalize to novel environments and goals found in SILG evaluation, we additionally fine-tune them via RL similar to~\citet{guo2019hybridRL}.
This method is listing as~\texttt{inverse} in experiment figures.
In addition to the data used by~\texttt{reward}~and~\methodnameshort,~\texttt{inverse}~uses additional rollouts to train the inverse dynamics model.

\subsection{Results and ablations}

\paragraph{\methodnameshort~consistently improves performance across environments.}
We evaluate on held-out environments across 4 random seeds for NetHack in Fig~\ref{fig:nethack}, Touchdown in Fig~\ref{fig:touchdown}, ALFWorld in Fig~\ref{fig:alfworld}, RTFM in Fig~\ref{fig:rtfm}, and Messenger in Fig~\ref{fig:messenger}.
\methodnameshort~obtains top performance compared to tabula-rasa policy learning with language descriptions, VAE pretraining, reward shaping using the dynamics model, and inverse reinforcement learning.
This is consistent across challenges in multi-step reasoning (RTFM), language-entity generalization (Messenger), large language action spaces (ALFWorld), large procedurally generated states (NetHack), and long natural language instructions with complex visual scenes (Touchdown).
We also ablate~\methodnameshort~by removing the initialization step (\texttt{LDD-init}) or the distillation step (\texttt{LDD-distill}).
On Messenger, methods that pretrain to initialize~(\methodnameshort,~\texttt{inverse},~\texttt{LDD-distill}) outperform those that do not~(\texttt{reward},~\texttt{LDD-init}).
On Touchdown, methods that distill or reward shape~(\methodnameshort,~\texttt{reward},~\texttt{LDD-init}) outperform those that do not.
Learning curves for each method across environments are shown in Appendix section~\ref{sec:learning_curves}.
\methodnameshort~converges faster and to a higher win rate than other methods, with the exception of Touchdown, where it achieves lower training but higher evaluation win-rate.

\begin{figure}
  \centering
  \includegraphics[width=0.7\linewidth]{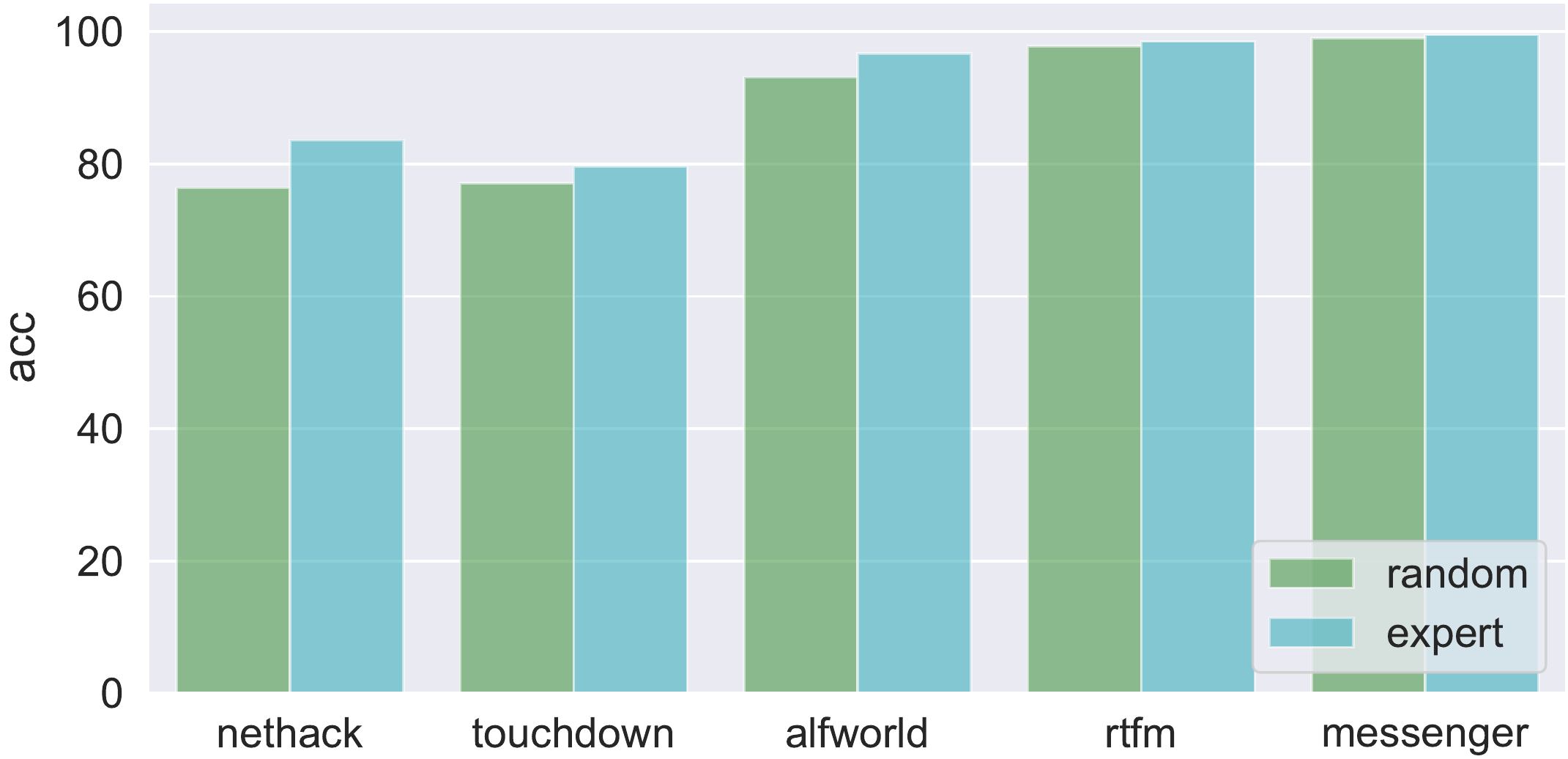}
  \caption{Dynamics model frame prediction accuracy using unlabeled expert demonstrations vs. rollouts from random policies. Training using expert demonstrations result in more accurate dynamics models, especially on complex environments that are difficult to explore via random policies.}
  \label{fig:expert_vs_rand_dm}
\end{figure}

\paragraph{Adding language descriptions improves performance.}
What is the role of language descriptions in pretraining and subsequent policy learning?
To answer this question, we ablate language descriptions by removing them from the environments in NetHack and SymTouchdown (the other environments are not solvable without descriptions because they describe the objective).
\texttt{no-msg} removes NetHack messages describing events near the agent (e.g.~\emph{kitten attacks the bat!}).
\texttt{no-msg,stats} additionally removes character state descriptions (e.g.~health, achievements, dungeon level).
\texttt{no-desc} removes Touchdown captions that describe objects locations scene relative to the agent (e.g.~\emph{on your left, there are many building, some people, and few cars}).
These variants differ only in the observation space used in dynamics modeling.
In fine-tuning, they receive the same observation space with language descriptions.
For both NetHack (Fig~\ref{fig:nethack}) and Touchdown (Fig~\ref{fig:touchdown}), removal of in-game messages, character state descriptions, and synthetic captions degrade performance.
Appendix Fig~\ref{fig:touchdown_dm}~further shows that adding language descriptions results in dynamics models with higher accuracy.
This suggests that modeling language descriptions in the observation space result in better initialization and distillation, which improve subsequent policy learning.

\paragraph{Expert demonstrations cover late-stage strategy that result in asymptotic gains.}
In complex environments, experts demonstrate late stage strategies difficult to explore via random sampling.
Because these stages are rarely reached, accurate dynamics models are especially helpful in providing signals when rewards are sparse.
Take NetHack as an example: unlike experts, non-expert policies never proceed to the deeper dungeons of the game.
A dynamics model trained on non-expert rollouts therefore struggles to generalize to unseen deeper dungeons.
As the agent learns and descends deeper in the dungeon, a dynamics model trained on non-expert demonstrations results in less distillation gains.
Fig~\ref{fig:expert_vs_rand_dm} compares dynamics modeling from observations using expert demonstrations vs. using rollouts from a random policy.
The evaluation is on a held-out set of expert demonstrations.
Across environments, training on expert demonstrations outperforms training on non-expert demonstrations.
This effect is less apparent on environments where random policies can (eventually) discover most of the state space (e.g, RTFM, Messenger) and more apparent on partially observed environments where only strategic expert policies can encounter rare states indicative of success (e.g.~long-term planning in NetHack and Touchdown, choosing from large language action space in ALFWorld).

\section{Conclusion}
While recent work showed that augmentation with language descriptions result in better policies, learning how to ground language descriptions to observations is difficult through naive RL with sparse, delayed rewards.
We proposed~\methodname, which pretrains a dynamics model using cheaply-obtained unlabeled demonstrations with language descriptions to initialize and distill into the policy learner.
On five tasks with language descriptions,~\methodnameshort~improved sample efficiency and resulted in better policies than RL from scratch, inverse RL, and expert reward shaping.
In addition, the benefit from initialization and distillation differ on an environment basis, but are complementary across environments.
Moreover, language descriptions improved initialization and distillation gains in policy learning.
Finally, learning to model dynamics with expert demonstrations was more effective than with non-expert rollouts.
A promising direction for future research is studying whether dynamics modeling with language descriptions is similarly effective in robotic control where naive RL can be prohibitively expensive, but unlabeled demonstrations with synthetic captions are cheap to obtain.

\begin{ack}
We are grateful to the anonymous reviewers for their helpful comments and suggestions.
Victor is supported in part by the ARO (AROW911NF-16-1-0121) and by the Apple AI/ML fellowship.
\end{ack}

\bibliography{myrefs}
\bibliographystyle{plainnat}

\clearpage

\section*{Checklist}

\begin{enumerate}

\item For all authors...
\begin{enumerate}
  \item Do the main claims made in the abstract and introduction accurately reflect the paper's contributions and scope?
  Yes
    \answerYes{}
  \item Did you describe the limitations of your work?
    \answerYes{See section~\ref{sec:limitation}}
  \item Did you discuss any potential negative societal impacts of your work?
    \answerYes{See section~\ref{sec:negative_societal_impact}}
  \item Have you read the ethics review guidelines and ensured that your paper conforms to them?
    \answerYes{}
\end{enumerate}

\item If you ran experiments...
\begin{enumerate}
  \item Did you include the code, data, and instructions needed to reproduce the main experimental results (either in the supplemental material or as a URL)?
    \answerYes{See section~\ref{sec:code_release}}
  \item Did you specify all the training details (e.g., data splits, hyperparameters, how they were chosen)?
    \answerYes{See section~\ref{sec:training}}
        \item Did you report error bars (e.g., with respect to the random seed after running experiments multiple times)?
    \answerYes{}
        \item Did you include the total amount of compute and the type of resources used (e.g., type of GPUs, internal cluster, or cloud provider)?
    \answerYes{See section~\ref{sec:compute}}
\end{enumerate}

\item If you are using existing assets (e.g., code, data, models) or curating/releasing new assets...
\begin{enumerate}
  \item If your work uses existing assets, did you cite the creators?
    \answerYes{}
  \item Did you mention the license of the assets?
    \answerYes{See section~\ref{sec:asset_license}}
  \item Did you include any new assets either in the supplemental material or as a URL?
    \answerNo{}
  \item Did you discuss whether and how consent was obtained from people whose data you're using/curating?
    \answerYes{See section~\ref{sec:asset_license}}
  \item Did you discuss whether the data you are using/curating contains personally identifiable information or offensive content?
    \answerYes{See section~\ref{sec:asset_license}}
\end{enumerate}

\end{enumerate}


\appendix
\section{Limitations}
\label{sec:limitation}
This work studies how language descriptions in unlabeled demonstrations benefit learning from observations.
The environments used in this work are simulations.
Despite variety across grounding challenges, performance on these environments do not necessarily transfer to other applications such as robotic control.
A promising direction for future work is to investigate whether dynamics modelling on language observations show similar benefits in other applications.

\section{Potential negative societal impacts}
\label{sec:negative_societal_impact}
The methodology in this work are based on reinforcement learning, which may learn uninterpretable policies that achieve the objective in surprising ways (e.g.~a robot that bumps along the cabinet while fetching dishes to clean).
Language-conditioned policies are a way of controlling how policies behave by adjusting the language (e.g.~instructions, in this case observations), however more research in this area is needed to develop methods that reliably understand and use language.

\section{Code release}
\label{sec:code_release}

The source code for our experiments is available at

\url{https://github.com/vzhong/language-dynamics-distillation}.

\section{Training details}
\label{sec:training}

We train and evaluate all methods on the SILG benchmark~\citep{zhong2021silg}, which comes with its own training and validation splits in terms of environment instances.
We make distinction for Nethack, where we train and evaluate on the more difficult Nethack Challenge~\citep{kuttler2020nethack}, and Messenger, where we train and evaluate on the second curriculum stage as opposed to the first curriculum stage.

The code bases for each environments are based on the following work
\begin{enumerate}
    \item Nethack: we use the~\href{Nethack Learning Environment code base}{https://github.com/facebookresearch/nle} and its hyperparameters with the~\texttt{human-monk} starting character. The demonstrations are 100k sampled~\texttt{ttyrec} screen recordings downloaded from~\href{alt.org}{nethack.alt.org}.
    \item RTFM: we use the~\href{SILG code base}{https://github.com/vzhong/silg} and its default hyperparameters. The demonstrations are 10k sampled trajectories from a converged agent released with SILG.
    \item Messenger: we use the~\href{EMMA code base}{https://github.com/ahjwang/messenger-emma} and its default hyperparameters on the second curriculum stage. The demonstrations are 10k sampled trajectories from a converged agent trained using the default settings in EMMA on stage 1, then adapted to stage 2 via curriculum learning.
    \item Touchdown: we use the~\href{SILG code base}{https://github.com/vzhong/silg} and its default hyperparameters. The demonstrations are the 6.5k human trajectories from the orinal Touchdown dataset~\citep{chen2018touchdown}.
    \item ALFWorld: we use the~\href{SILG code base}{https://github.com/vzhong/silg} and its default hyperparameters. The demonstrations are the 21k full state planner trajectories from the original ALFRED dataset~\citep{shridhar2020alfred}.
\end{enumerate}

The dynamics models trained on these demonstrations are re-used for the reward-shaping method.
To collect data for inverse dynamics modelling, we train a policy using the same hyperparameters for each environment for 10k episodes, then sample 10k episodes from the resulting policy.
The sampled 10k episodes are used to learn a inverse dynamics model where two consecutive frames are used as the input and the inverse model predicts the action that took place between the frames.
This inverse model is then used to predict actions on the demonstrations.
The (pseudo-labeled) demonstrations are then used for imitation learning.
The hyperparameters of the imitation learner is the same as those of the LDD experiments.
This imitation learned model is then fined-tuned with RL.

Code for running the environment is anonymously submitted in the link in Section~\ref{sec:code_release}.
Hyperparameters for our experiments are obtained from~\citet{zhong2021silg} for RTFM, ALFWorld, and Touchdown; ~\citet{hanjie2021grounding} for Messenger, and~\citet{kuttler2020nethack} for NetHack.
They are reproduced in Table~\ref{tab:hyperparam} for convenience.

\begin{table}[th]
    \centering
    \begin{tabular}{cccccc}
        \toprule
        Name & RTFM & Messenger & NetHack & ALFWorld & Touchdown \\
        \midrule
        Base model & SIR & EMMA & ChaoticDwarf & SIR & SIR \\
        Embedding size & 100 & 256 & 128 & 100 & 30 \\
        RNN size & 200 & & 128 & 200 & 100 \\
        Final repr size & 400 & 256 & 128 & 400 & 200 \\
        Num FiLM$^2$ layers & 5 & & & 5 & 3\\
        Entropy cost & 0.05 & 0.05 & 0.001 & 0.05 & 0.05 \\
        Baseline cost & 0.5 & 0.5 & 0.25 & 0.5 & 0.5 \\
        Optimizer & RMSProp & Adam & Adam & RMSProp & RMSProp \\
        Learning rate & 5e-5 & 1e-4 & 1e-4 & 5e-4 & 5e-4 \\
        Optim epsilon & 0.01 & 1e-6 & 1e-6 & 0.01 & 0.01 \\
        RMSProp alpha & 0.99 &  &  & 0.99 & 0.99 \\
        Adam beta1 &  & 0.99 & 0.99 &  &  \\
        Adam beta2 &  & 0.999 & 0.999 &  &  \\
        Num actors & 30 & 30 & 128 & 30 & 8 \\
        Learner batch size & 24 & 24 & 128 & 10 & 3\\
        Learner threads & 4 & 4 & 4 & 4 & 4\\
        Unroll length & 80 & & 64 & 80 & 64\\
        \bottomrule
    \end{tabular}
    \caption{Hyperparameter settings. The base models SIR, EMMA, and ChaoticDwarf are respectively described in~\citet{zhong2021silg},~\citet{hanjie2021grounding}, and~\citet{kuttler2020nethack}.}
    \label{tab:hyperparam}
\end{table}

\section{Compute resources}
\label{sec:compute}
We use a slurm cluster to train models.
Each machine is equipped with a NVIDIA GPU with at least 16GB RAM and 20 CPU cores.
Each run typically last 3 days, with the exception of ALFWorld (10 days) and Touchdown (6 days).
Across 5 environments, we run 4 methods and 2 ablations for a total of 6 experiments.
We additionally run 2 more language ablations experiments for Nethack and 1 more for Touchdown.
Each experiment consists of 4 random seeds for a total of $(5 \times 6 + 3) \times 4 = 132$ runs.
For the policy learning stage, our resource usage are on the order of $132 \times 10 \times 24 = 32k$ GPU hours or $132 \times 10 \times 24 \times 20 = 634k$ CPU hours.
These experiments compose the bulk of our resource usage.

For dynamics pretraining, each run takes approximately 2 days of 1 GPU and 4 CPU.
We re-use the trained dynamics model for reward shaping experiments.
Inverse-dynamics modelling additionally require 1 day of pretraining an initial non-expert policy, generating rollouts from said policy, learning a inverse-dynamics model, and annotating unlabeled demonstrations with the inverse-dynamics model.

\section{Asset and license}
\label{sec:asset_license}
We distribute this work under the MIT license.
The dataset we use are publically available and distributed as a part of the SILG benchmark~\citep{zhong2021silg}.
There are no personally identifying information in the assets we use.
SILG is distributed under a MIT license.
The included environments are licensed as follows:
\begin{enumerate}
    \item NetHack: \href{https://github.com/facebookresearch/nle/blob/master/LICENSE}{NetHack General Public License}
    \item Touchdown: \href{https://github.com/lil-lab/touchdown/blob/master/LICENSE.txt}{Creative Commons Attribution 4.0 International}
    \item ALFWorld: \href{https://github.com/alfworld/alfworld/blob/master/LICENSE}{MIT}
    \item RTFM: \href{https://github.com/facebookresearch/RTFM/blob/master/LICENSE}{Attribution-NonCommercial 4.0 International}
    \item Messenger: \href{https://github.com/ahjwang/messenger-emma/blob/master/LICENSE}{MIT}
\end{enumerate}

\clearpage

\section{Learning curves}
\label{sec:learning_curves}

Note that NetHack does not have a held-out evaluation set of environments.

\begin{figure}[!htb]
    \centering
    \includegraphics[width=0.85\linewidth]{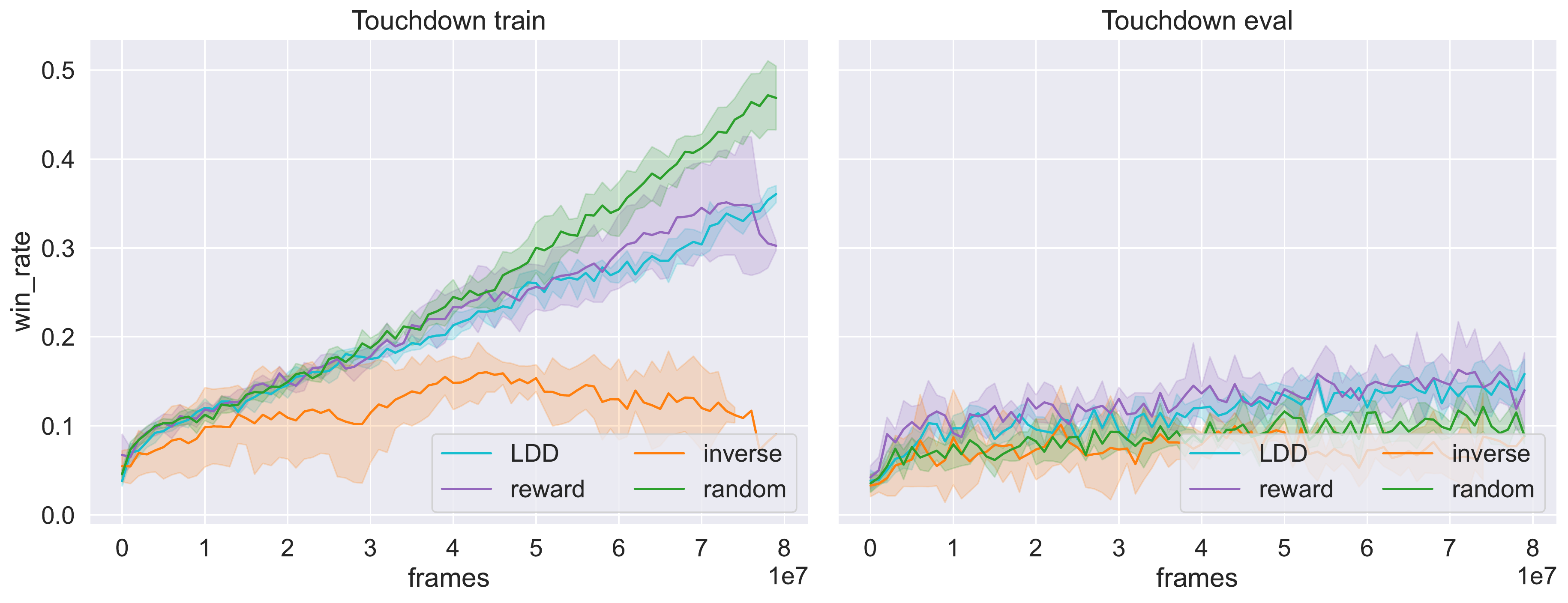}
    \caption{Touchdown learning curves}
    \label{fig:touchdown_lc}
\end{figure}

\begin{figure}[!htb]
    \centering
    \includegraphics[width=0.85\linewidth]{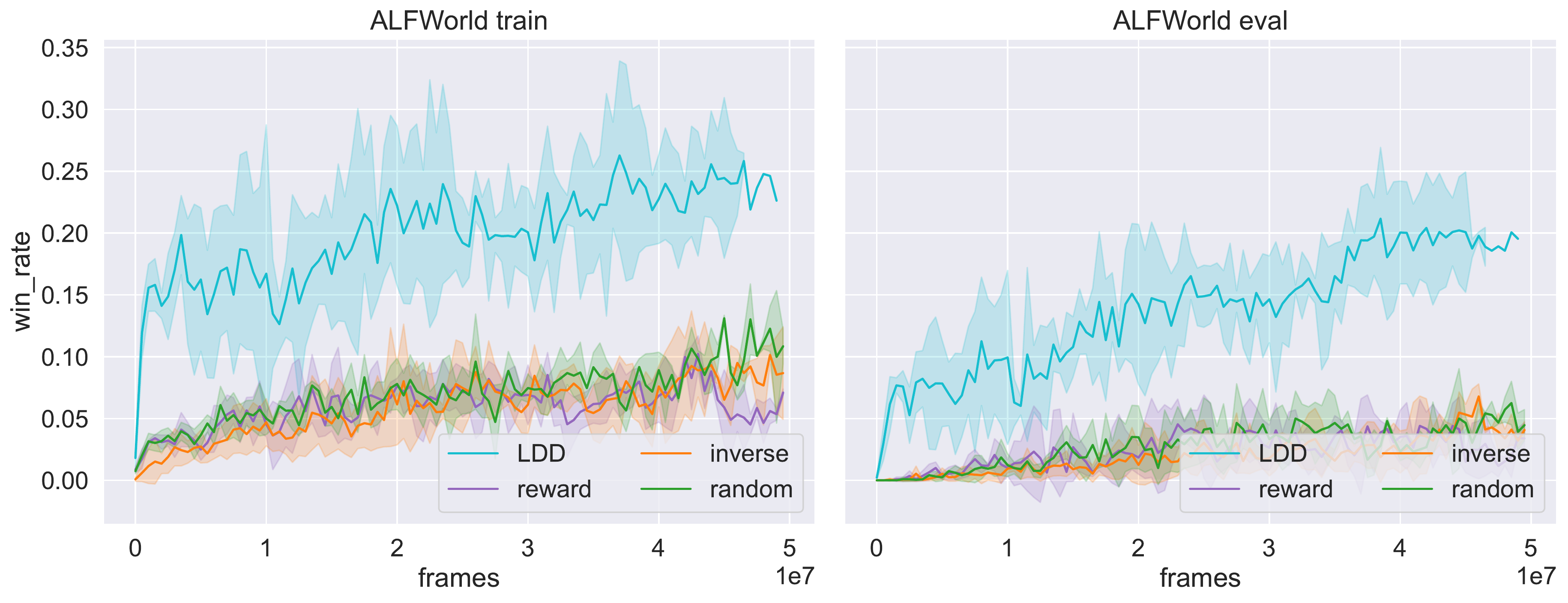}
    \caption{ALFWorld learning curves}
    \label{fig:alfworld_lc}
\end{figure}

\begin{figure}[!htb]
    \centering
    \includegraphics[width=0.85\linewidth]{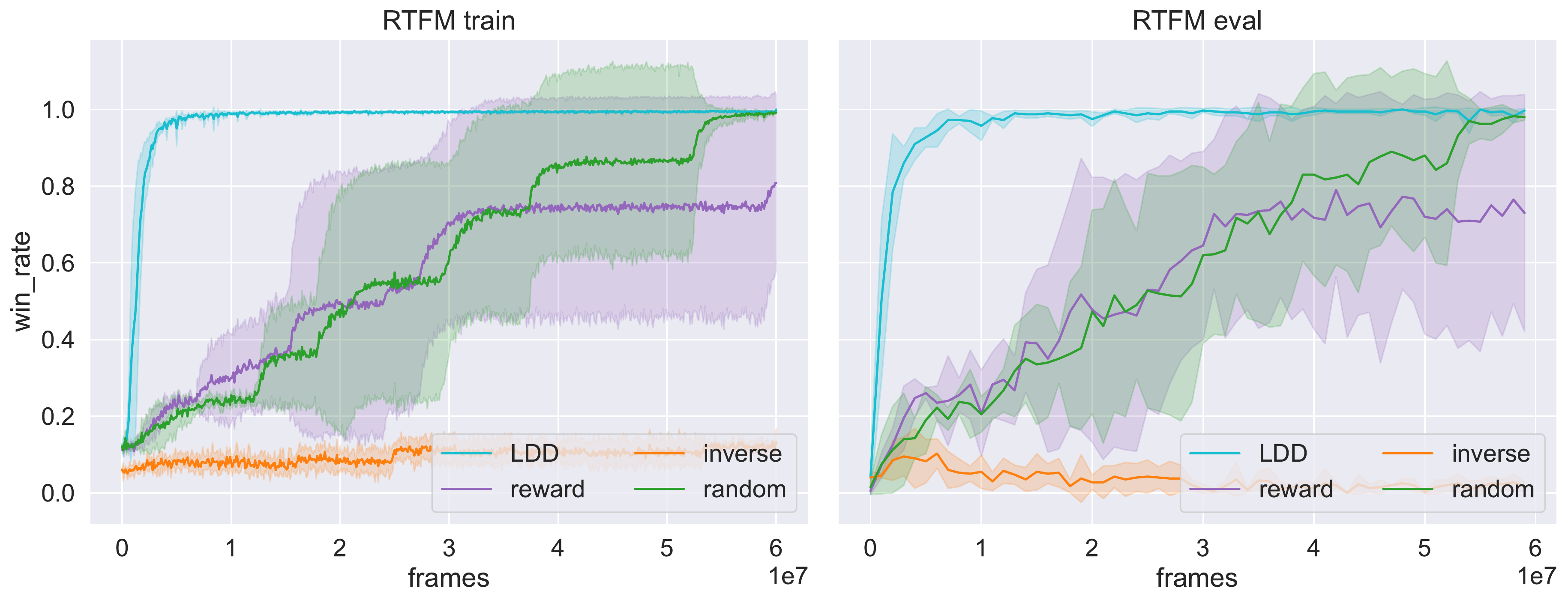}
    \caption{RTFM learning curves}
    \label{fig:rtfm_lc}
\end{figure}

\begin{figure}[!htb]
    \centering
    \includegraphics[width=0.85\linewidth]{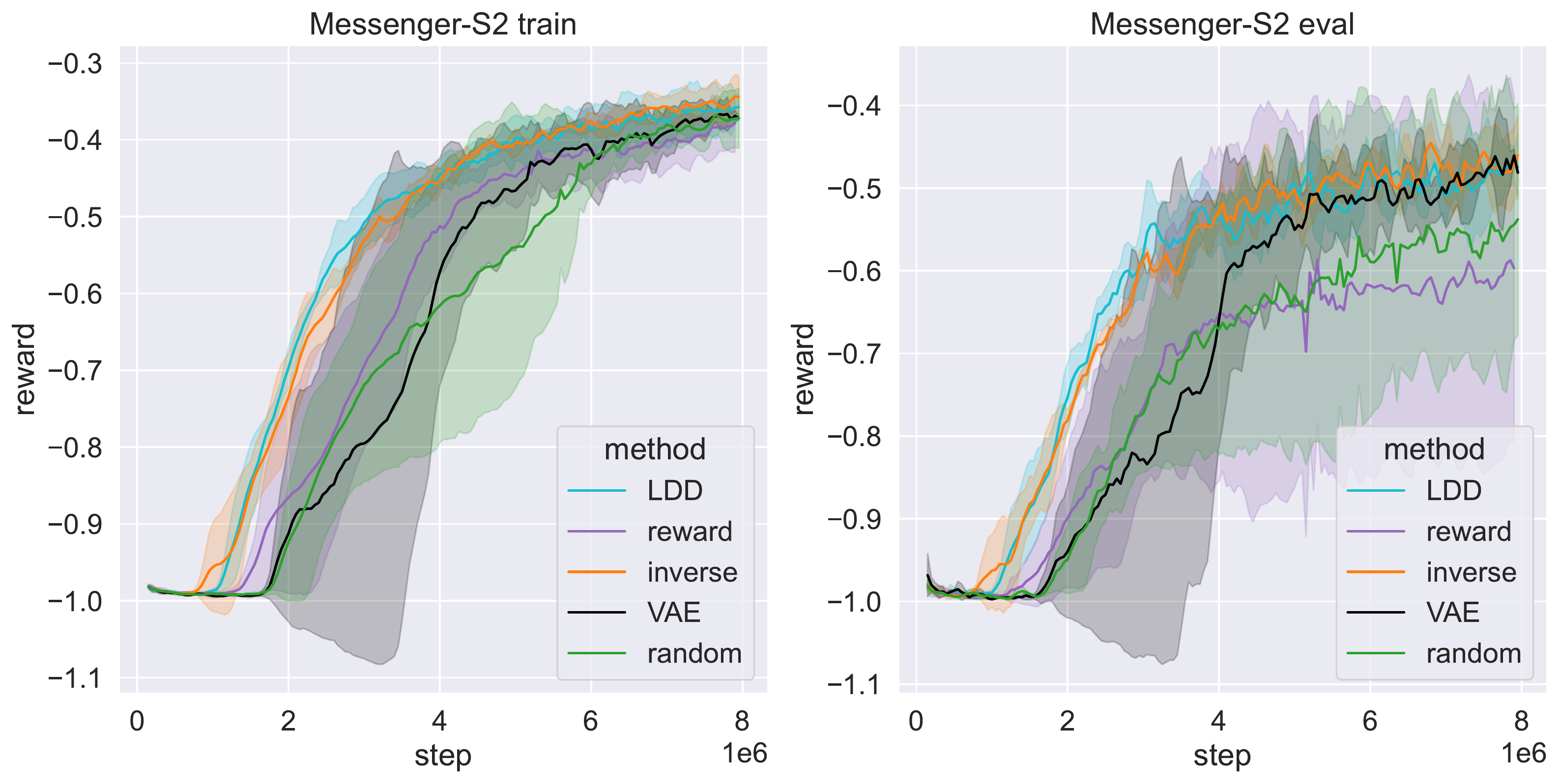}
    \caption{Messenger learning curves}
    \label{fig:messenger_lc}
\end{figure}

\newpage

\section{Dynamics modelling with vs. without language}

\begin{figure}[h]
    \centering
    \includegraphics[width=\linewidth]{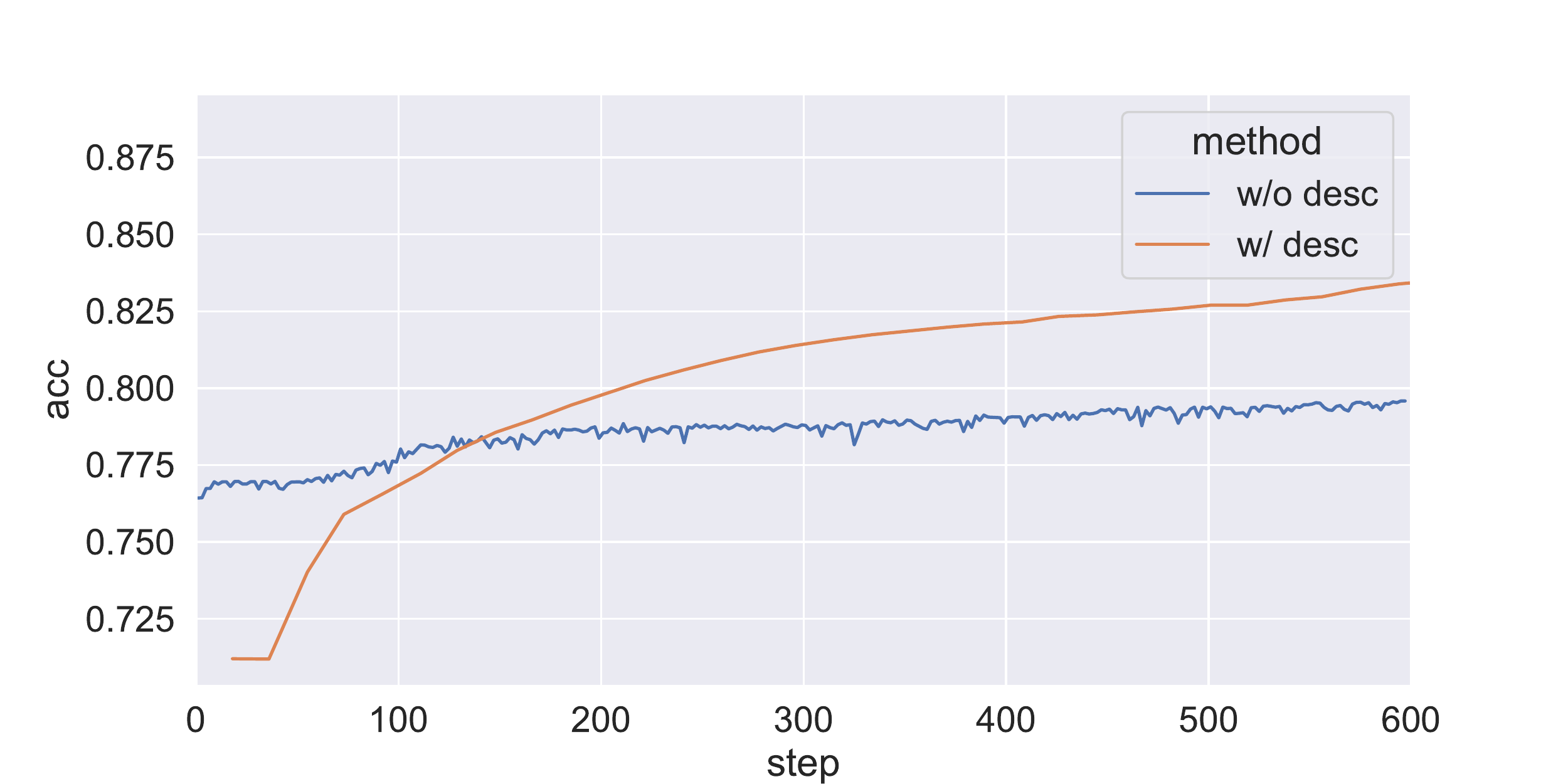}
    \caption{Dynamics modelling pixel-wise accuracy for SymTouchdown with vs. without language description inputs.}
    \label{fig:touchdown_dm}
\end{figure}

\end{document}